\documentclass[runningheads, a4paper]{llncs}
\usepackage[T1]{fontenc}
\usepackage{amsmath}
\usepackage{graphicx}
\usepackage{verbatim}
\usepackage[ruled,vlined,noend]{algorithm2e}
\usepackage[noend]{algpseudocode}
\usepackage{caption}
\usepackage{subcaption}

\usepackage{fancyhdr}
\usepackage{ragged2e}
\fancypagestyle{custom_notice}
{
    \setlength{\headheight}{32pt}
    \addtolength{\topmargin}{-5pt}
    \lhead{\fbox{\begin{minipage}{\textwidth}
    \noindent\footnotesize\justifying\copyright 2025 This paper has been accepted to the 19th International Symposium on Experimental Robotics, ISER-2025. Personal use of this material is permitted.\end{minipage}}}
    \cfoot{} % removes page number
}

\begin{document}
%
%Practical Considerations for In-Hand Friction Estimation
\title{Friction Estimation for In-Hand Planar Motion 
\thanks{This work was funded by the Wallenberg AI, Autonomous Systems and Software Program (WASP) funded by the Knut and Alice Wallenberg Foundation.\\
Website: \url{https://wasp-sweden.org/}}
%\begin{abstract}
%This paper focuses on online contact property estimation under planar motions for in-hand sliding manipulations, where unmodeled fast dynamics may occur. The paper examines the practical challenges of friction estimation in real-world systems and presents strategies for handling unmodeled dynamics to ensure reliable estimation. Key contributions include an online estimator for the real-time estimation of static friction $mu_s$ and Coulomb friction $c$ as well as radius $r$ of a rim contact approximation. The estimator is first tested in simulations and subsequently on a sensorized gripper that can sense both wrench and velocity. Simulation results are analyzed and a simple filtering heuristic is proposed to improve the performance of the estimator for real-world experiments.
%\end{abstract}
}
\author{Gabriel Arslan Waltersson \inst{1}\orcidID{0000-0001-6775-0584} \and
Yiannis Karayiannidis\inst{2}\thanks{The author is a member of the ELLIIT Strategic Research Area at Lund University.}\orcidID{0000-0001-5129-342X} }
\authorrunning{G. A. Waltersson, Y. Karayiannidis}
\institute{Department of Electrical Engineering, Chalmers University of Technology, SE-41296 Gothenburg, Sweden \\ \email{gabwal@chalmers.se}  \and
Department of Automatic Control, Lund University, SE-22100 Lund, Sweden\\ \email{yiannis@control.lth.se}
}
\maketitle              % typeset the header of the contribution
\thispagestyle{custom_notice}

\begin{abstract}
This paper presents a method for online estimation of contact properties during in-hand sliding manipulation with a parallel gripper. We estimate the static and Coulomb friction as well as the contact radius from tactile measurements of contact forces and sliding velocities. The method is validated in both simulation and real-world experiments. 
Furthermore, we propose a heuristic to deal with fast slip-stick dynamics which can adversely affect the estimation. 
%The abstract should briefly summarize the contents of the paper in 150--250 words.

\end{abstract}

\section{Introduction and Related Work}

Humans can quickly infer surface properties when handling unfamiliar objects—an ability that enables us to assess grasp stability, minimize effort, and perform fine manipulation tasks such as controlled sliding.  Building on our previous work on in-hand sliding \cite{In_hand_gripper} and friction modeling for simulation \cite{friction_modeling}, this paper focuses instead on online friction estimation for in-hand sliding manipulation.
%Humans can quickly infer surface properties when handling unfamiliar objects. This ability allows us to assess grasp stability, minimize effort, and perform in-hand sliding manipulations. Building on our previous work on in-hand sliding manipulation \cite{In_hand_gripper} and friction modeling for simulation \cite{friction_modeling}, this paper focuses instead on online friction estimation for in-hand sliding manipulation. 
The sensorized gripper designed in \cite{In_hand_gripper} is able to measure both forces and sliding velocities at the contact surfaces. We present an estimator for the real-time estimation of static friction $\mu_s$ and Coulomb friction $\mu_c$ as well as radius $r$ of a rim contact approximation \cite{friction_modeling}. The paper examines the practical challenges of friction estimation in real-world systems and presents strategies for handling unmodeled dynamics to ensure reliable estimation. 

Existing review papers cover tactile sensing in robotic grippers \cite{Dahiya2010tactile}, \cite{reviewTactileSensorsZhanat2015}, and friction estimation \cite{Chen2018FrictionEstimation}. While prior work primarily focuses on friction estimation for grasping and slip prevention, we target in-hand sliding manipulation. The friction at the contact surface is influenced by numerous factors, including surface roughness \cite{Dahiya2010tactile}, sliding velocity, contact area, temperature, humidity, age of contact, and applied force rate \cite{Chen2018FrictionEstimation}. Given friction's sensitivity to typically unmodeled and uncontrolled factors, real-time online estimation becomes essential for reliable performance.

Friction can be estimated using incipient slip or gross slippage. Okatani \emph{et. al} \cite{Okatani2016StaticFric} proposed a tactile sensor that estimates static friction when making contact with a surface by measuring the incipient slip due to sensor deformation, allowing estimation of static friction before gross slippage.  In \cite{Maria2015est_fric}, friction coefficients were estimated during an initial exploration phase to prevent slippage. A combined vision-haptic approach was introduced in \cite{Le2021Prob_friction_est}, where the material was segmented via vision and guided exploration of haptic friction estimation.

The relationship between static and torsional friction during the onset of slip was explored in \cite{Dahmen2005Macroscopic}. During sliding, torque and tangential friction coupling can be described by the limit surface \cite{goyal1989limit}, \cite{goyal1991planar}. Shi \emph{et. al} \cite{Shi2017Dynamic_in_hand} used the limit surface to model gripper contact with a flat object for dynamic repositioning. The estimation of contact and friction is also relevant for pushing tasks, where \cite{Lee2024Pushing} used an unscented Kalman filter to estimate the model parameters, including friction. Like in-hand manipulation, pushing tasks involve planar motions, but are typically assumed quasi-static with constant normal force.

%This paper focuses on online contact property estimation under planar motions for in-hand manipulation, where unmodeled fast dynamics may occur. Key contributions include an online estimator for $\mu_s$, $\mu_c$, and $r$,  analysis of coupling in the estimation of $\mu_c$ and $r$, validation through simulations and experiments, and insights into practical challenges with solutions for reliable estimation.

This paper focuses on online contact property estimation under planar motions for in-hand manipulation, where unmodeled fast dynamics may occur. The key contributions are: 
\begin{enumerate} 
    \item An online contact property estimator for $\mu_s$, $\mu_c$, and $r$, analysis of coupling in the estimation of $\mu_c$ and $r$.
    \item Validation through simulations and real-world experiments. 
    \item Insights into practical challenges with solutions for reliable estimation. 
\end{enumerate}

The remainder of the paper is structured as follows: Section \ref{sec:Technical} introduces the estimator. Section \ref{sec:experiments} introduces the simulations and real-world experiments. The results and experimental insights are presented in Section \ref{sec:main_insights} together with heuristic extensions of the estimator. Finally, Section \ref{sec:conclusions} concludes the paper.

\section{Technical Approach}\label{sec:Technical} 

This section formulates a recursive least-squares  (RLS) estimator for friction coefficient and contact radius estimation under planar motions. RLS is chosen for its simplicity and suitability in scenarios where a comprehensive system model is unavailable. Instead of relying on complex models, we employ simpler approximations within an online estimator. Although the joint estimation of friction coefficients and contact radius is inherently nonlinear—due to their coupling within the friction limit surface—we show that the coupling within the estimator is stable.

\begin{figure}[ht]
    \vspace{-1.0cm}
    \centering
    \begin{subfigure}[b]{0.35\textwidth}
        \includegraphics[width=\textwidth]{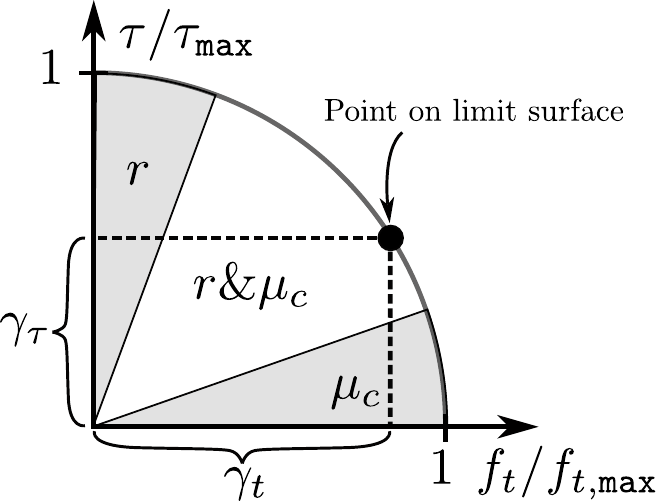}
        \caption{Illustration of normalized limit surface and regions of estimation.}
        \label{fig:limit_surf}
    \end{subfigure}
    \hfill
    \begin{subfigure}[b]{0.35\textwidth}
        \includegraphics[width=\textwidth]{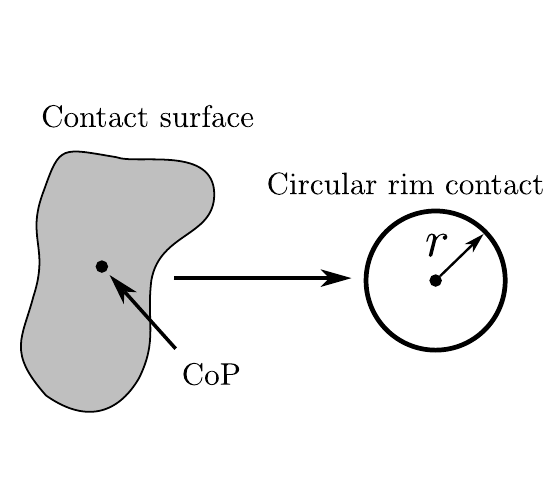}
        \caption{Rim contact approximation of a contact surface.}
        \label{fig:rim_contact}
    \end{subfigure}
    \caption{Models of contact properties.}
    \label{fig:contact_surface}
    \vspace{-0.8cm}
\end{figure}

Assuming an equivalent rim contact surface of radius $r$ \cite{friction_modeling}, see Fig. \ref{fig:rim_contact}, we can use a twist with uniform units $\left [v_x\;v_y\;r\omega\right ]^T$ to describe the planar in-hand object motion where $v_x$, $v_x$ are the coordinates of the linear velocity of the center of pressure (CoP) and $\omega$ the angular velocity. Let $v_t = \sqrt{v_x^2 + x_y^2}$ and $v = \sqrt{v_t^2 + r^2\omega^2}$ be the Euclidean norms of linear velocity and scaled twist respectively.

Under planar motion, torque and tangential friction forces are coupled; for Coulomb friction, the coupling is described by the limit surface, see Fig. \ref{fig:limit_surf}. Our estimator approximates the limit surface using an ellipsoid assuming a rim-contact (Fig. \ref{fig:rim_contact}) for which the tangential force $f_t$ and  torque $\tau$ are given by:
\begin{align}
           f_t &= (\gamma_t f_n) \mu_c \textrm{ with } \gamma_t = \frac{v_t}{v} \label{eq:est_mu}\\
            \tau &= (\gamma_\tau \mu_c f_n) r \textrm{ with }\gamma_{\tau} = \frac{|r \omega|}{v}\label{eq:est_r}
\end{align}
which holds when $v > \epsilon_v$, ensuring that the effect of $\mu_s$ is negligible. The estimation of $\mu_s$, $\mu_c$, and $r$ is divided into two regimes: 
\begin{description}
\item[Stick-slip regime:] $\mu_s$ is estimated from stick-slip events. When such an event is detected via velocity measurements, $\mu_s$ can be inferred from the force data.
\item[Motion regime:] $\mu_c$ and $r$ are estimated from the motion regime ($v > \epsilon_v$).
\end{description}
During the motion regime, assuming dry friction, the coefficients $\mu_c$ and $r$ can be estimated from \eqref{eq:est_mu} and \eqref{eq:est_r}.
The estimation of $r$ and $\mu_c$ is bootstrapped using prior estimates of $r$ and the most recent estimates of $\mu_c$. The coupling due to bootstrapping is analyzed at the end of the section. 
%Assuming dry friction for our in-hand manipulation scenario, allows $\mu_c$ and $r$ to be estimated from \eqref{eq:est_mu} and \eqref{eq:est_r}, and $\mu_c$ in \eqref{eq:est_r} is estimated from \eqref{eq:est_mu}. 
Different regions of the limit surface contain different amounts of information about $\mu_c$ or $r$. Specifically:
\begin{itemize}
    \item $\mu_c$ is more identifiable when tangential velocity dominates ($v_t >> r\omega$),
    \item  $r$ is better identified when rotational velocity dominates ($r\omega  >> v_t$).
\end{itemize}
To quantify this, we use the coefficients $\gamma_t$ and $\gamma_\tau$ as indicators of the observability of $\mu_c$ and $r$, respectively, see see Fig. \ref{fig:limit_surf}. Estimates are updated only when:
\begin{itemize}
    \item $\gamma_t > \epsilon_t$ for $\mu_c$,
    \item $\gamma_\tau > \epsilon_\tau$ for $r$.
\end{itemize}
Both parameters are estimated using RLS with an initial covariance matrix $P_0$ and forgetting factor $\lambda$. The estimator is active only when contact is detected, i.e., when the normal force $f_n \geq \epsilon_{f_n}$.  

During the stick-slip regime,  Algorithm \ref{alg:est_mu_s} is deployed to estimate $\mu_s$. The algorithm saves $\frac{f_t}{\gamma_t f_n}$ in a buffer of length $n_b$ and when slippage occurs the mean of the highest $n_a$ values is used as an estimate of $\mu_s$. 

%The overall estimation process for both regimes is described in Algorithm. \ref{alg:est_mu_s}.

\begin{algorithm}[ht]
\SetAlgoLined
buffer $\leftarrow$ Buffer($n_b$)\;
Initialize: $v_z \leftarrow 1$; $n_a \leftarrow 2$; $\hat{\mu}_s \leftarrow \hat{\mu}_s(t_0)$\;

\For{each time step}{
    \uIf{$v == 0$}{
        $\gamma_t \leftarrow f_t / \left(\sqrt{f_x^2 + f_y^2 + (r\tau)^2}  + 10^{-10}\right)$ \tcp*{From force}
    }
    \Else{
        $\gamma_t \leftarrow v_t / \left\| \mathbf{v} \right\|_2$ \tcp*{From velocity}
    }

    \uIf{$\gamma_t > \epsilon_t$}{
        $\hat{\bar{\mu}}_s \leftarrow f_t / (\gamma_t f_n)$\;
        buffer.append($\hat{\bar{\mu}}_s$)\;
    }
    \Else{
        \textbf{publish} $\mu_s$; \textbf{continue};
    }

    \If{$(v_z == 1) \wedge (v_t > v_s)$ \textbf{or} $(v_z == 0) \wedge (v_t == 0)$}{
        $\hat{\mu}\_s \leftarrow \text{avg}(\textit{get\_n\_largest}(n_a, \text{buffer}))$\;
        $v_z \leftarrow 1 - v_z$\;
        buffer.clear()\;
    }
    \textbf{publish} $\hat{\mu}_s$; \tcp*[f]{Output current $\mu_s$ for external use}  
}

\caption{\small Estimation of $\mu_s$\label{alg:est_mu_s}}
\end{algorithm}

The estimation of $\mu_c$ and $r$ are nonlinearly coupled through $\gamma_t$ and $\gamma_\tau$ and to show the stability of the coupling during $\gamma_t > \epsilon_t$ and $\gamma_\tau > \epsilon_\tau$, we can define a update mapping:
\begin{equation}
    T : \mu_c \rightarrow \frac{f_t}{f_n \gamma_t(\mu_c)}
\end{equation}
where we write $\gamma_t$, $r$, and $\gamma_\tau$ as function of $\mu_c$:
\begin{equation*}
     \gamma_t(\mu_c) = \frac{v_t}{\sqrt{v_t^2 + (r(\mu_c)\omega)^2}}, \quad r(\mu_c) = \frac{\tau}{\gamma_\tau(\mu_c)\mu_c f_n}, \quad \gamma_\tau(\mu_c) = \frac{r(\mu_c)\omega}{\sqrt{v_t^2 + (r(\mu_c) \omega)^2}}
\end{equation*}
and the true friction coefficient $\mu_c$ satisfies the fixed-point condition $\mu_c = T(\mu_c)$. By analyzing how an added error $\hat{\mu}_c = \mu_c + \delta$ will propagate through the estimator, an change in $\mu_c$ (i.e. using $\hat{\mu}_c$), results in the gradients:
\begin{equation}
    \frac{\partial r}{\partial \mu_c} < 0, \quad \frac{\partial \gamma_t}{\partial r} < 0, \quad \frac{\partial \gamma_t}{\partial \mu_c} = \frac{\partial \gamma_t}{\partial r}\frac{\partial r}{\partial \mu_c} > 0.
\end{equation}
Hence $\frac{\partial}{\partial \mu_c} T'(\mu_c) < 0$, and under the assumption of $|\frac{\partial}{\partial \mu_c} T'(\mu_c)| < 1$, then Banach fixed-point theorem shows that starting with an error $\delta$ the estimation will self correct over the next iterations.

\section{Experiments}\label{sec:experiments}
\begin{figure}[ht]
    \centering

    \begin{subfigure}[b]{0.35\textwidth}
        \includegraphics[width=\textwidth]{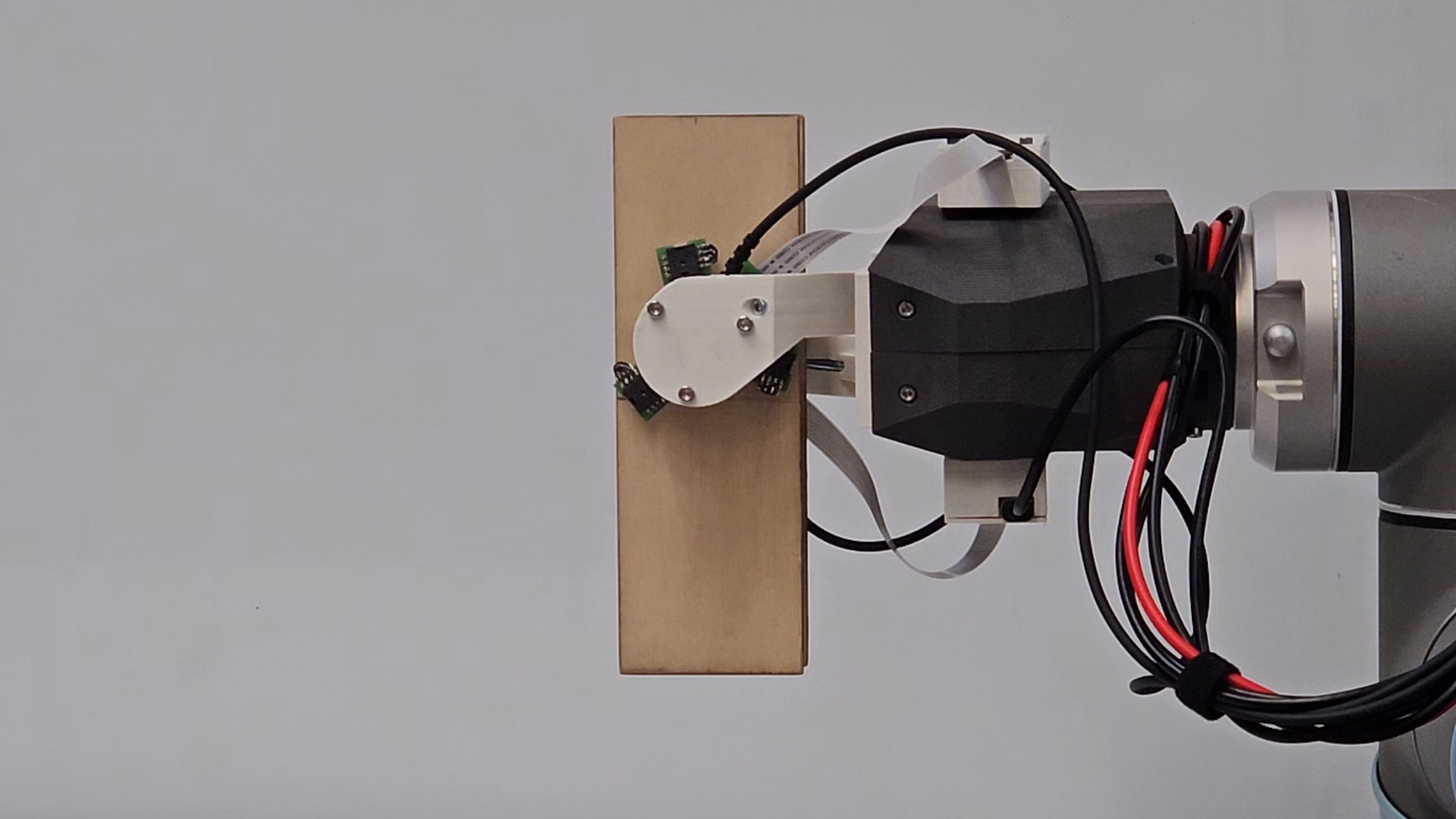}
        \caption{The gripper and sensors mounted on a robotic arm.}
        \label{fig:gripper}
    \end{subfigure}
    \hfill
    \begin{subfigure}[b]{0.291\textwidth}
        \includegraphics[width=\textwidth]{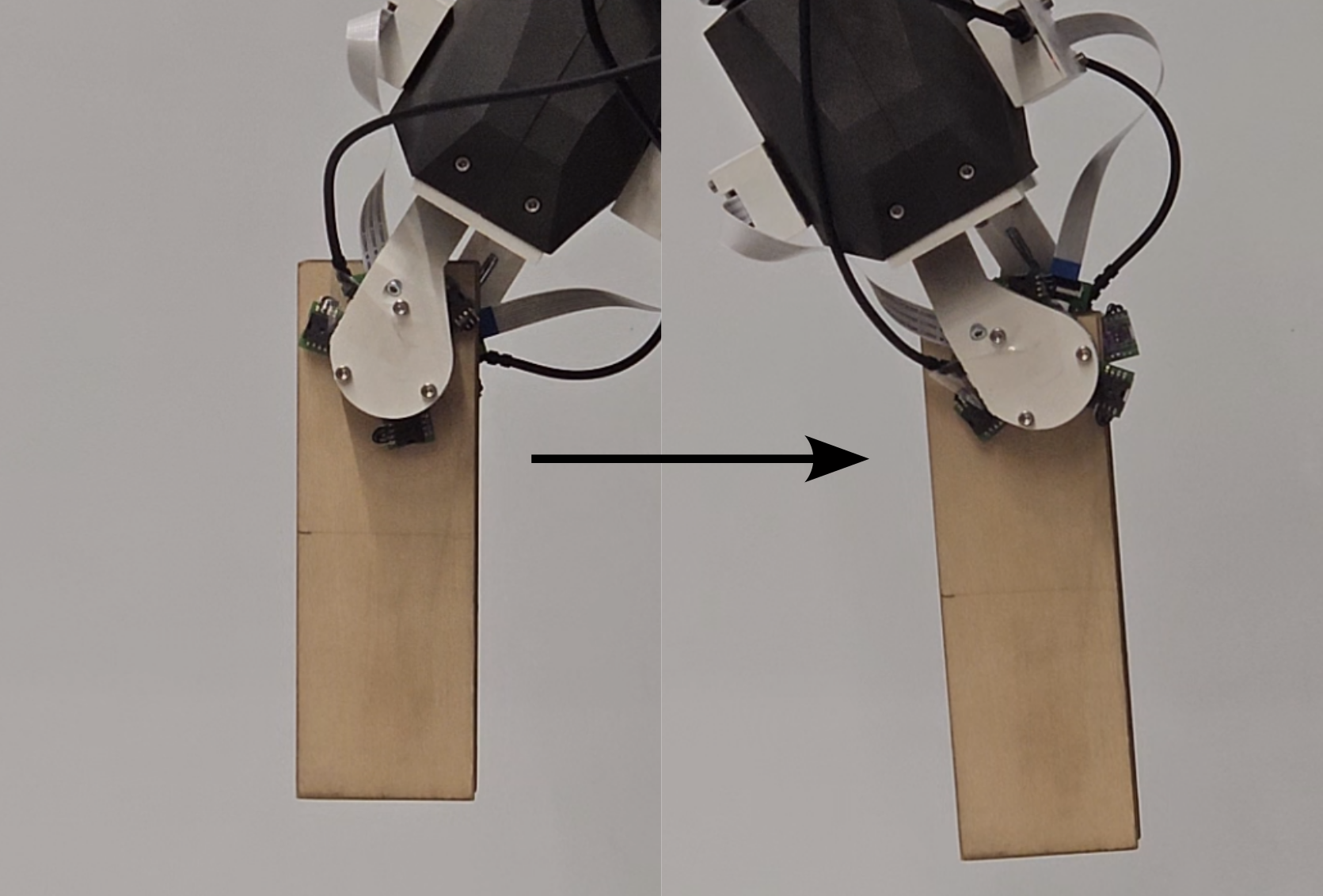}
        \caption{The object under planar motion.}
        \label{fig:gripper_hinge}
    \end{subfigure}
    \hfill
    \begin{subfigure}[b]{0.323\textwidth}
        \includegraphics[width=\textwidth]{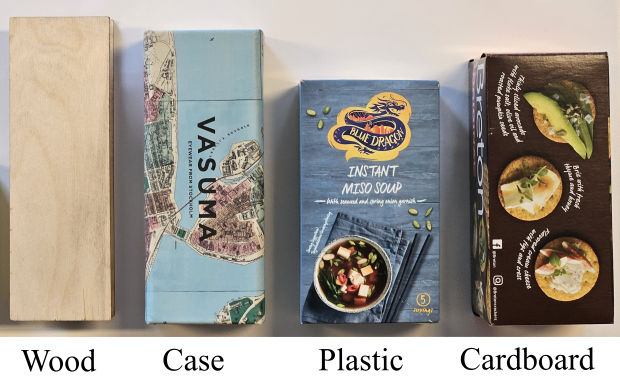}
        \caption{The objects tested.}
        \label{fig:objecs}
    \end{subfigure}

    \caption{Experimental setup.}
    \label{fig:experiment_setup}
    \vspace{-0.6cm}
\end{figure}

The experiment consists of both simulation experiments to validate that the estimator follows the ground truth and real-world experiments, where an heuristic is added to improve the real-world performance. This paper uses the gripper, controllers and sensors developed in \cite{In_hand_gripper}, see Fig. \ref{fig:gripper}. The gripper allows for rapid force control and is equipped with tactile force and velocity sensors on each finger. The  ATI mini40 F/T (Force/Torque) sensors are capable of measuring 6 DoF force wrench at 1000 Hz, and a custom relative velocity senors capable of measuring $v_x$, $v_y$, and $\omega$ of the slippage at 120 Hz. We tested the estimator on data collected from experiments with this setup and with four different objects with different materials, see Fig. \ref{fig:objecs}. 

The estimator is first tested in simulation, by simulating a disc with uniform pressure following a pre-defined velocity profile (see Fig. \ref{fig:Simulation_results}) using the simulator developed in \cite{friction_modeling}, which is capable of modeling the stick-slip phenomenon for surfaces under planar motions. During the simulation the true friction and contact radius is changed to test how well the estimator follows the ground truth. The estimator is then tested on real-world data under linear slippage, for the linear slippage a total of 160 trials were gathered, consisting of 4 object, 4 different time and distance combinations, and 10 trials for each, more details in \cite{In_hand_gripper}. The estimator is first tested without any added heuristics, the results are analyzed and a simple filtering heuristics is proposed to improve the performance of the estimator (discussed further in Section \ref{sec:main_insights}). Finally, the estimator is tested under planar motion, see Fig. \ref{fig:gripper_hinge}, with a total of 40 trials. In the planar case, there is both linear and rotational slippage.  

\section{Main Experimental Insights}\label{sec:main_insights}

\begin{figure}
    \centering
    \captionsetup{font=small}
    \includegraphics[width=0.99\linewidth]{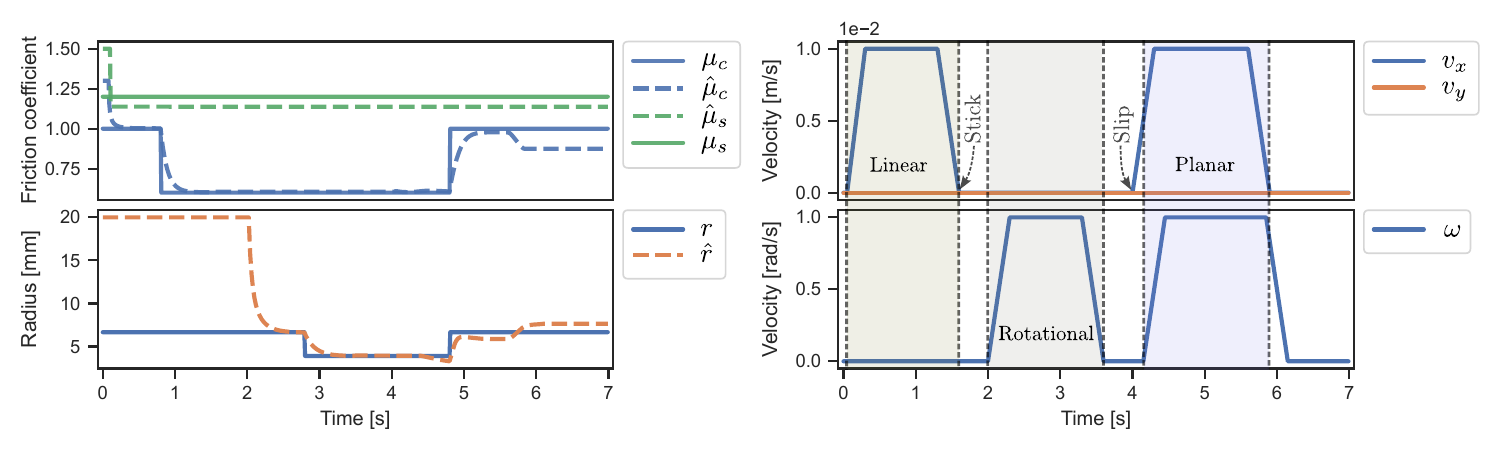}
    \caption{Simulation with changing radius and friction coefficient under linear, rotational and planar motion, and where the stick-slip events occur in the transitions.}
    \label{fig:Simulation_results}
    \vspace{-0.7cm}
\end{figure}

The results of the simulation experiment are presented in Fig. \ref{fig:Simulation_results}, and the parameters used across all experiments are listed in Table \ref{tab:estimator_parameters}, where $(t_0)$ indicates the initial value at time $t=0$. As shown in Fig. \ref{fig:Simulation_results}, the estimator successfully identifies $\mu_s$ from the stick-slip events, though a slight bias is introduced due to averaging over the  $n_a$ highest values. The estimator successfully identifies $\mu_c$ during linear motion, $r$ during rotational motion, and estimates both parameters simultaneously during planar motion. Moreover, the estimator is capable of adapting to time-varying friction coefficients and contact radius, even when both change concurrently during planar sliding. It should be noted that the friction model used to simulate the data is not the same model as in the estimator, as the estimator used an ellipsoid approximation of the limit surface. The discrepancy is highlighted in \cite{friction_modeling} and can be clearly seen in the results from $t>6$ s. However, the ellipsoid approximation offers simplicity and reduces the problem to estimating only the contact radius instead of the pressure distribution.   

\begin{table}[pbb]
    \centering
    \footnotesize
    \captionsetup{font=small}
    \begin{tabular}{c|c|c|c|c|c|c}
        \hline
        $\hat{\mu}_c(t_0) = 1.3$ & $\hat{r}(t_0) = 0.02$ [m] & $\epsilon_\tau = 0.3$ & $\epsilon_t = 0.3$ & $\epsilon_v = 1.5\mathrm{e}{-3}$ & $n_a=2$ & $\Delta t = 0.05$ [s] \\ \hline
        $\hat{\mu}_s(t_0) = 1.5$ & $\epsilon_{f_n} = 0.2$ [N] &  $P_0 = 1$ & $\lambda = 0.98$ & $n_b = 16$ & $\epsilon_\delta = 150$ [N/s] & $\delta t = 1/120 [s]$\\ \hline
    \end{tabular}
    \caption{Estimator parameters.}
    \label{tab:estimator_parameters}
     \vspace{-0.5cm}
\end{table}

\begin{figure}[ht]
    \centering
    \captionsetup{font=small}
    \includegraphics[width=1.0\linewidth]{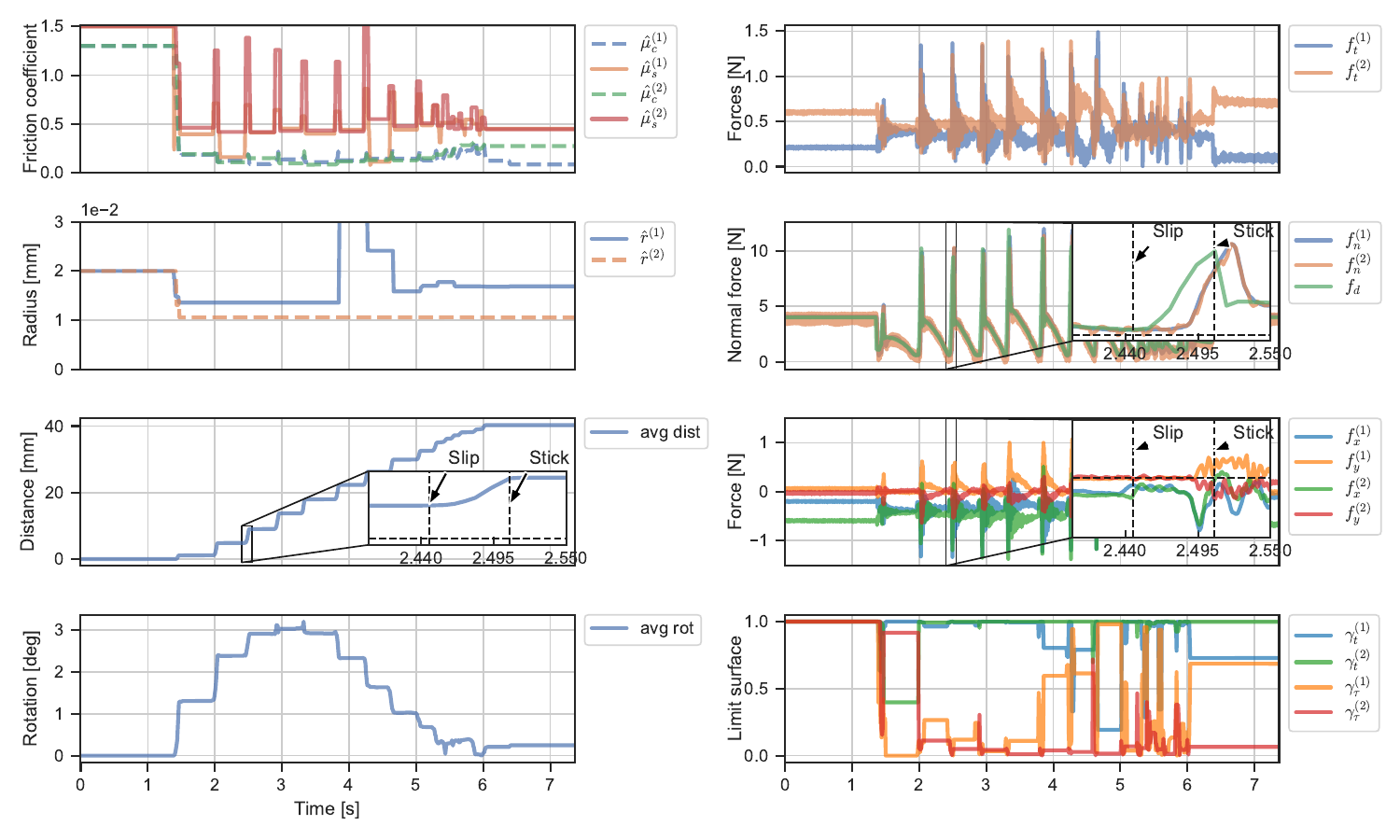}
    \caption{Estimation results and forces of each sensor holding a plastic covered object that linearly slips 40 mm under 5 s (no heuristic).}
    \label{fig:no_heuristics}
    \vspace{-0.7cm}
\end{figure}

Estimation in real-world is rarely as trivial as in simulation. A sample result from applying the estimator on real-world data can be seen in Fig. \ref{fig:no_heuristics}. The estimated $\mu_c$ is lower than expected from the forces and fluctuating. There are multiple causes; from Fig. \ref{fig:no_heuristics} it can be seen that there is a significant slip-stick effect and that the slip dynamics are very fast. The time duration from slipping to sticking is for some objects as low as 0.05 s. Analyzing the forces, it can be seen that there are interactions between both of the fingers, and that during stiction due to rapidly changing $f_n$ there are oscillating dynamics that are faster than what the velocity sensors can measure. The oscillations in the forces are not due to measurement noise, but can be attributed to very fast dynamics. Furthermore, the timing between the signals being crucial for estimation discourages averaging filters before the estimator. The outcome is that $\mu_c$ is typically underestimated when no additional heuristic is added. The results for the linear slippage can be seen in Table \ref{tab:no_h},  where $\bar{\cdot}$ is the mean, $\sigma_{\bar{\cdot}}$ is the standard deviation of the mean between experiments, and $\bar{\sigma}_{\cdot}$ is the mean standard deviation within an experiment. From \eqref{eq:est_r}, it follows that the contact radius $r$ will be overestimated if $\mu_c$ is underestimated due to the measurements, which can be observed in Table \ref{tab:no_h} when no heuristics is added. The outer radius of the contact surface is 15 mm and it would be expected that a perfect uniform pressure would have an effective radius of 10 mm. 

\begin{table}[pbb]
    \centering
    \footnotesize
    \captionsetup{font=small}
    \resizebox{\columnwidth}{!}{%
    \begin{tabular}{c||c|c|c||c|c|c||c|c|c}
        Material & $\bar{\mu}_c$ & $\sigma_{\bar{\mu}_c}$ & $\bar{\sigma}_{c}$  & $\bar{\mu}_s$ & $\sigma_{\bar{\mu}_s}$ & $\bar{\sigma}_s$ &  $\bar{r}$ & $\sigma_{\bar{r}}$ & $\bar{\sigma}_r$ \\ \hline
        Wood  no heuristics & 0.2249 & 0.019 & 0.0448  & 0.4032 & 0.0404 & 0.1162 & 0.0192 & 0.006 & 0.0028 \\ 
        Wood with heuristics  & 0.3623 & 0.0341 & 0.042 & 0.5155 & 0.0977 & 0.1052 & 0.0124 & 0.0041 & 0.0012 \\ \hline
        Case  no heuristics & 0.3039 & 0.0846 & 0.0809& 0.5673 & 0.0481 & 0.1275 & 0.0202 & 0.0059 & 0.0046 \\ 
        Case with heuristics & 0.4208 & 0.0412 & 0.0533 & 0.606 & 0.0614 & 0.1227 & 0.0165 & 0.0044 & 0.0024 \\ \hline
        Plastic  no heuristics & 0.1101 & 0.0177 & 0.0299 & 0.4343 & 0.0389 & 0.0979 & 0.0194 & 0.0048 & 0.0037 \\ 
        Plastic with heuristics & 0.3646 & 0.0344 & 0.0275& 0.5705 & 0.0706 & 0.0866 & 0.0168 & 0.0023 & 0.0003  \\ \hline
        Cardboard  no heuristics & 0.2334 & 0.0425 & 0.0843 & 0.519 & 0.0392 & 0.1119 & 0.0253 & 0.0075 & 0.0042 \\ 
        Cardboard with heuristics & 0.4039 & 0.0325 & 0.0478  & 0.5819 & 0.0507 & 0.0939 & 0.0184 & 0.0048 & 0.0011 \\ \hline
    \end{tabular}
    }
    \caption{Estimation of $\mu_c$, $\mu_s$, and $r$ under linear motion.}
    \label{tab:no_h}
    \vspace{-0.7cm}
\end{table}

\begin{figure}[ht]
    \centering
    \includegraphics[width=0.65\linewidth]{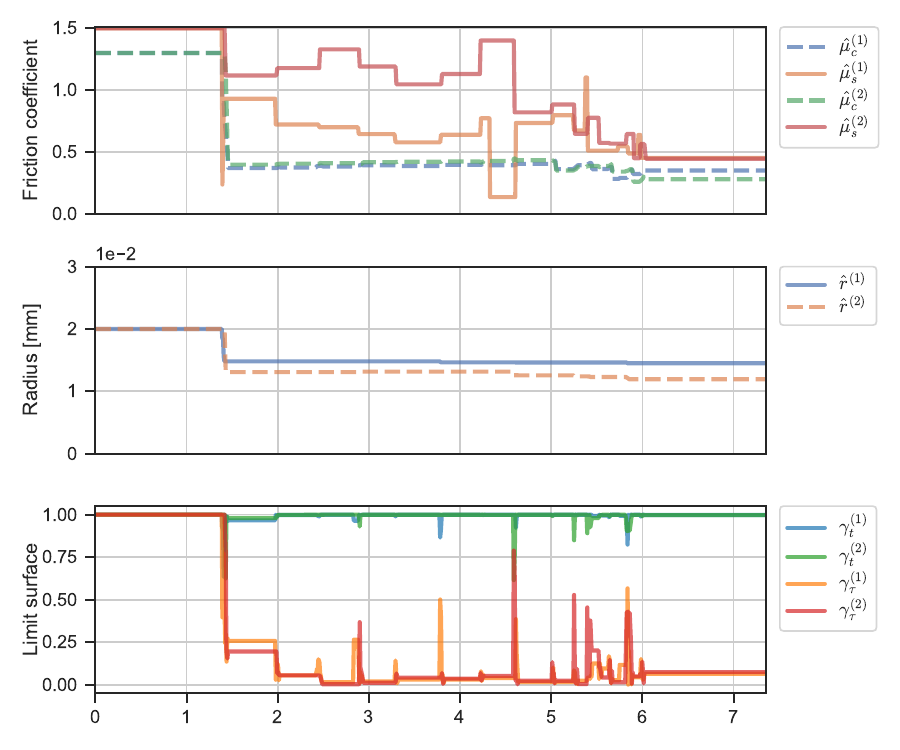}
    \caption{Estimation results and forces of each sensor holding a plastic covered object that linearly slips 40 mm under 5 s (with heuristic).}
    \label{fig:dfn_heuristics}
    \vspace{-0.5cm}
\end{figure}

To improve estimation, we introduce a heuristic based on the condition: 
\begin{equation}\label{eq:heuristic}
    \frac{f_n(t) - f_n(t-\delta t)}{\delta t} > \epsilon_{\delta}
\end{equation}
%$\frac{f_n(t) - f_n(t-\delta t)}{\delta t} > \epsilon_{\delta}$
as a filter to temporarily halt the estimation over a time interval of $\Delta t$ enables the estimation to take place in regions where the dynamics are less adverse. The estimator is run with the added heuristic on the same data as in Fig. \ref{fig:no_heuristics}, and the updated results are shown in Fig. \ref{fig:dfn_heuristics}. Here it can be seen that the estimated $\mu_c$ values are higher and more stable, it can also be seen that both fingers of the gripper estimate similar $\mu_c$ values.
The improved results of the heuristic, for all materials, can be seen in Table \ref{tab:no_h}. The results show that with the heuristic the estimation of $\mu_c$ is consistent with the forces measured and less fluctuating. From Fig. \ref{fig:no_heuristics} and  \ref{fig:dfn_heuristics}, it can be observed that the estimator rapidly updates and converges to a new value of $\mu_c$ when slippage occurs.

The estimator is tested under planar motion, where both rotational and linear motion occur at the same time. This is achieved by letting the gripper hold the object vertically and then rotating the gripper around the object, see Fig. \ref{fig:gripper_hinge}. The results for each material can be seen in Table \ref{tab:est}, and shows that the estimation outcome is similar to the linear case. It is noteworthy that, as the normal force changes less during the planar experiment, the heuristic has less effect on the estimation, which can be observed in Table \ref{tab:est}. 

\begin{figure}[ht]
    \centering
    \includegraphics[width=1.0\linewidth]{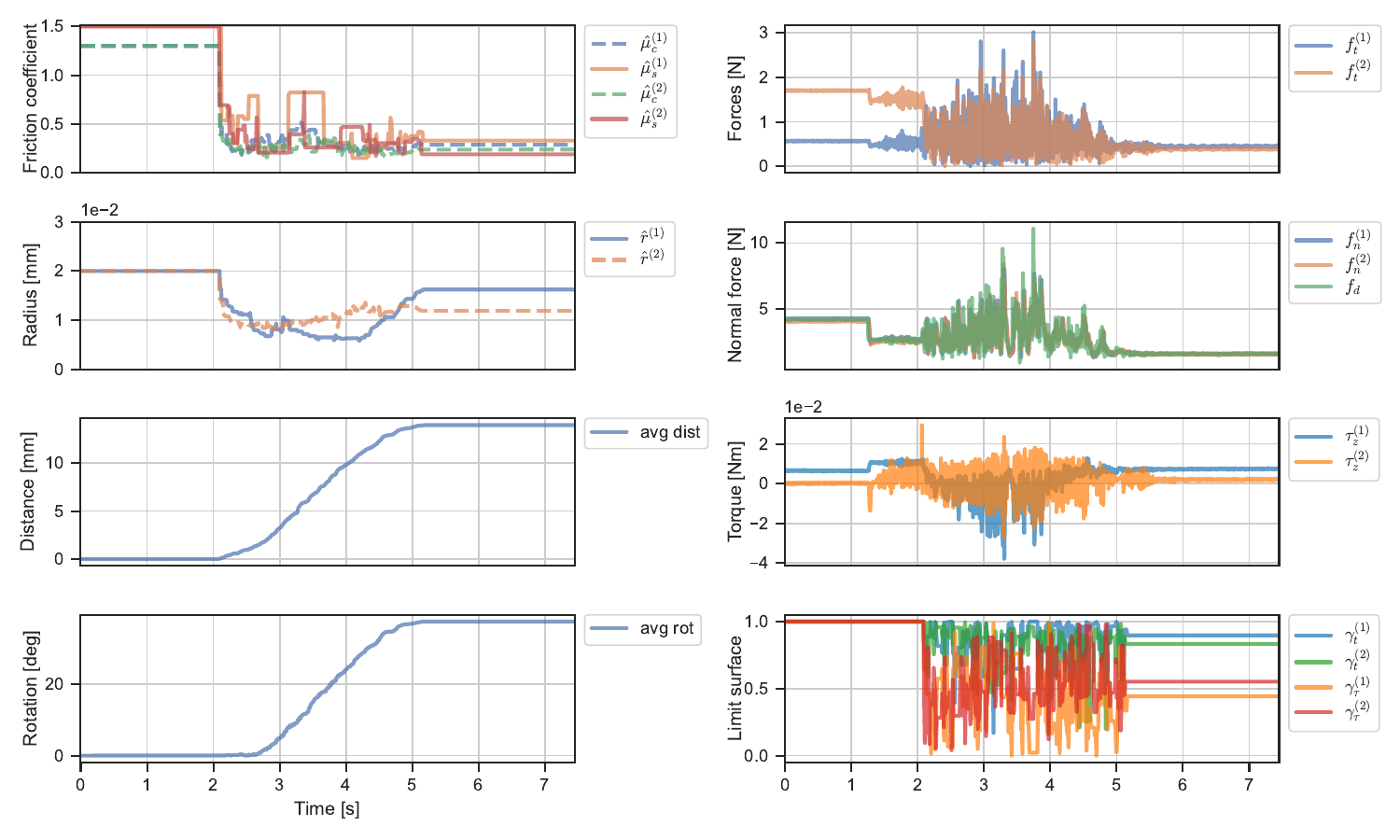}
    \caption{Estimation results and forces from each sensor holding a plastic covered object that both slips and rotates (planar motion with the heuristic).}
    \label{fig:dfn_heuristics_planar}
    \vspace{-0.5cm}
\end{figure}

\begin{table}[pbb]
    \centering
    \footnotesize
    \captionsetup{font=small}
    \resizebox{\columnwidth}{!}{%
    \begin{tabular}{c||c|c|c||c|c|c||c|c|c}
        Material & $\bar{\mu}_c$ & $\sigma_{\bar{\mu}_c}$ & $\bar{\sigma}_{c}$  & $\bar{\mu}_s$ & $\sigma_{\bar{\mu}_s}$ & $\bar{\sigma}_s$ &  $\bar{r}$ & $\sigma_{\bar{r}}$ & $\bar{\sigma}_r$ \\ \hline
        Wood no heuristics & 0.3123 & 0.0163 & 0.0747 & 0.4153 & 0.0512 & 0.1424 & 0.0147 & 0.0026 & 0.0052 \\
        Wood with heuristics & 0.325 & 0.014 & 0.0741 & 0.431 & 0.0956 & 0.1178 & 0.0157 & 0.0011 & 0.0057 \\ \hline
        Case no heuristics & 0.393 & 0.0305 & 0.0763 & 0.5415 & 0.1041 & 0.1372 & 0.0149 & 0.0018 & 0.0059 \\
        Case with heuristics & 0.422 & 0.0299 & 0.062 & 0.5183 & 0.1077 & 0.1191 & 0.0147 & 0.0019 & 0.006 \\ \hline
        Plastic no heuristics & 0.2744 & 0.0211 & 0.0729 & 0.4169 & 0.0518 & 0.1657 & 0.0151 & 0.0035 & 0.0066 \\
        Plastic with heuristics & 0.3024 & 0.0224 & 0.0714 & 0.422 & 0.0328 & 0.1399 & 0.0134 & 0.0044 & 0.0067 \\ \hline
        Cardboard no heuristics & 0.413 & 0.0447 & 0.09 & 0.5541 & 0.0623 & 0.168 & 0.0161 & 0.006 & 0.0068 \\
        Cardboard with heuristics & 0.4204 & 0.0407 & 0.0879& 0.548 & 0.0667 & 0.1646 & 0.0158 & 0.0055 & 0.0065 \\ \hline
    \end{tabular}
    }
    \caption{Estimation of $\mu_c$, $\mu_s$, and $r$ under planar motion.}
    \label{tab:est}
\end{table}

During linear motion (e.g., Fig. \ref{fig:dfn_heuristics}), the ratio $\gamma_t$ dominates over $\gamma_\tau$, and the estimator primarily updates $\mu_c$ during slippage. In contrast, under planar motion (Fig. \ref{fig:dfn_heuristics_planar}), $\gamma_t$ and $\gamma_\tau$ are closer to each other, enabling simultaneous estimation of both $\mu_c$ and the contact radius $r$. While the heuristic improves the estimation in both cases (linear and planar), the estimated contact radius tends to be overestimated in both cases compared to the ideal scenario. Several factors contribute to this overestimation. The contact pressure distribution is rarely uniform, finger compliance and material deformation can shift the CoP, and these effects are not explicitly modeled. Additionally, the estimation method assumes an ellipsoidal approximation of the limit surface. Since the true limit surface depends on the unmeasured pressure distribution, this approximation introduces local inaccuracies that can bias the results. Nonetheless, the estimator remains capable of jointly estimating friction parameters and contact radius during planar motion.

The heuristic also improves the estimation of static friction $\mu_s$. As shown in Fig. \ref{fig:no_heuristics}, during linear slippage, the estimated $\mu_s$ is often lower during the striction phase than during slip, likely due to rapidly varying normal forces. To mitigate this, the same heuristic from \eqref{eq:heuristic} is applied for $\mu_s$ in Alg. \ref{alg:est_mu_s} to skip updates to $\mu_s$  when the condition is met.  The effect is evident in Fig. \ref{fig:dfn_heuristics}, where the $\mu_s$ estimate does not drop during each stick event.  However, since $\mu_s$ is inferred from fewer observations and exhibits greater stochasticity, its estimation remains less consistent than that of $\mu_c$.

\section{Conclusions}\label{sec:conclusions}

In this paper, we present a method for online estimation of contact properties for in-hand manipulation. The estimator does not rely on specific exploration procedures, except that slippage has to occur. The method is evaluated in both simulation and on real-world data. Additional challenges arising in real-world settings are discussed, and a heuristic is introduced to improve performance under these conditions. Future work will focus on jointly advancing both the estimator and the sensor design to enhance contact property estimation and object manipulation capabilities. The estimator is planed to be integrated into a framework for slip-aware in-hand manipulation with parallel grippers, extending the capabilities of robotic systems.

\bibliographystyle{splncs04}
\bibliography{references.bib}

\end{document}